\documentclass[review,5p,twocolumn]{elsarticle}
\usepackage{hyperref}
\usepackage{amsmath}
\usepackage{amssymb}

\journal{Journal of \LaTeX\ Templates}

\begin{document}

\begin{frontmatter}

\title{Dynamic Local Feature Aggregation for Learning on Point Clouds}


 \author[mymainaddress]{Zihao Li}
\ead{pride_19@163.com}

 \author[mymainaddress]{Pan Gao}
\ead{Pan.Gao@nuaa.edu.cn}

 \author[mysecondaryaddress]{Hui Yuan}
\ead{huiyuan@sdu.edu.cn}

 \author[mythirdaddress]{Ran Wei}
\ead{ 115946873@qq.com}


 \address[mymainaddress]{Nanjing University of Aeronautics and Astronautics, Nanjing ,China}
\address[mysecondaryaddress]{Shandong University, Jinan, China}
\address[mythirdaddress]{Science and Technology on Electro-optic Control Laboratory, Luoyang, China}

\begin{abstract}
Existing point cloud learning methods aggregate features from neighbouring points relying on constructing graph in the spatial domain, which results in feature update for each point based on spatially-fixed neighbours throughout layers. In this paper, we propose a dynamic feature aggregation (DFA) method that can transfer information by constructing local graphs in the feature domain without spatial constraints. By finding k-nearest neighbors in the feature domain, we perform relative position encoding and semantic feature encoding to explore latent position and feature similarity information, respectively, so that rich local features can be learned. At the same time, we also learn low-dimensional global features from the original point cloud for enhancing feature representation. Between DFA layers, we dynamically update the constructed local graph structure, so that we can learn richer information, which greatly improves adaptability and efficiency. We demonstrate the superiority of our method by conducting extensive experiments on point cloud classification and segmentation tasks. Implementation code is available: \href{https://github.com/jiamang/DFA}{https://github.com/jiamang/DFA}.
\end{abstract}


\end{frontmatter}


\section{Introduction}

The collection of points that express the spatial distribution and surface features of the target is called point cloud data, which  represents the 3D target in an unstructured form.  The point cloud obtained by combining the laser principle and the photography principle mainly contains three-dimensional position coordinates (X, Y, Z), laser reflection intensity and color information (R, G, B).  Common point cloud data formats include RGB-D dual-modality format and Point Cloud  space format. RGB-D dual-modality data records the color  information and depth information of the surface of the target object.  The Point Cloud space format records three-dimensional coordinates of the sampling points on the surface of the object, reflecting the spatial contour information.

Learning features from point clouds often requires a lot of advanced processing.  Traditional methods proposed to solve these problems include capturing the geometric characteristics of point clouds by using the hand-crafted features \cite{biasotti2016recent}. With the breakthrough of convolution neural network and deep learning, significantly better performance is achieved in various tasks of point cloud processing. However, standard deep neural network needs normative input data, but the point cloud data does not need to be irregular, and operations such as translation and rotation will not change its own nature. Some methods consider  converting to a normative 3D grid and then send the grid into the network for training, but it will cause additional memory occupation and information loss. Pointnet proposed by \cite{qi2017pointnet} creates a precedent for learning and processing directly on the original point cloud, where the multi-layer perceptron  is applied to each point. 

However, since Pointnet \cite{qi2017pointnet} cannot capture the contextual information, many recent studies have introduced different  modules to learn more abundant local structures, which can be divided into the following categories: 1) Feature update based on constructing graph structure \cite{wang2018local}\cite{shen2018mining}\cite{wang2019dynamic}\cite{wang2019graph}\cite{liu2019dynamic}; 2) Feature pooling based on neighboring points \cite{li2018so}\cite{mnih2014neural}\cite{huang2018recurrent}\cite{zhang2019shellnet}\cite{zhao2019pointweb}; 3) Convolution based on a series of kernels \cite{su2018splatnet}\cite{hua2018pointwise}\cite{wu2019pointconv}\cite{lan2019modeling}\cite{komarichev2019cnn}\cite{wu2019pointconv}\cite{mao2019interpolated}\cite{thomas2019kpconv}; 4) Learning  based on attention mechanism \cite{paigwar2019attentional}\cite{xie2018attentional}\cite{zhang2019pcan}\cite{yang2019modeling}. These methods have achieved good results in  classification and segmentation, but the construction of local feature learners and calculation of attention weight  have very expensive computing cost and memory occupation. In addition, the feature extractors proposed by some methods are not efficient enough, and there are many parts worth improving.

The goal of this paper is to design an efficient local feature extractor without adding much complexity, and then use the learned efficient features to represent objects, which will improve the point cloud classification and segmentation tasks. So we propose a dynamic feature aggregation (DFA) module, which extracts and learns latent features by finding k-nearest neighbors in the feature domain, encoding location information and semantic feature information simultaneously, and concatenating these two parts. In the classification and segmentation task, this module is stacked to extract rich local features. Using the network structure like Pointnet \cite{qi2017pointnet}, we extract low-dimensional global features from the initial point cloud, and then concatenate them with local features extracted by multiple DFAs. Finally, high-dimensional global features are obtained for classification and segmentation. For segmentation, we concatenate the high-dimensional global features again with local features, and perform the MLP operation to predict the category of each point.

In general, we design an efficient local feature extractor that utilizes multi-level and multi-source features to effectively characterize objects. Multi-level features are reflected in that by stacking several layers of DFA, we can gradually obtain deeper contextual features. Multi-source features are reflected in that we combine multiple types of features of location information, feature differences, features themselves, and low-dimensional global features to perform deeper and higher-dimensional feature learning. In order to test its efficiency, we have done relevant tests on the ModelNet40 \cite{wu20153d}, shapeNet \cite{yi2016scalable} and S3DIS \cite{armeni20163d} datasets. Furthermore, we also do many visualization results and ablation experiments.
Our main contributions are summarized as follows:
\begin{itemize}
    \item We propose a new operation DFA, which finds k-nearest neighbors in the feature domain to construct a local graph structure for feature aggregation at each time. The graph between DFA layers is dynamically updated, which is more adaptable.
    \item In each DFA layer, we can learn rich latent position and feature difference information through proposed relative position encoding and semantic feature encoding, respectively. To the best of our knowledge, simultaneously aggregating the relative position and feature information in the feature domain has not been studied before. 
    \item  We make full use of the learned local features and low-dimensional global features for point cloud classification and segmentation tasks, and test on benchmark datasets with outstanding quantitative and qualitative results.
\end{itemize}

\section{Related work}
\subsection{Voxel-based Network.}
 Converting point cloud data into regular voxel structure can preserve and express spatial distribution. In 2016, Qi \emph{et al.} \cite{qi2016volumetric} improved voxel CNN and proposed two different voxel CNN network structures. Afterwards, Tchapmi \emph{et al.} \cite{tchapmi2017segcloud} jointly proposed segcloud based on voxel-based 3D full convolution neural network and point based conditional random field. Wang \emph{et al.} \cite{wang2017cnn} proposed O-CNN. Its core idea is to use octree to represent 3D shapes, and only the sparse octree occupied by the shape boundary is subject to CNN operation. In order to effectively encode the distribution of voxel midpoint, Meng \emph{et al.} \cite{meng2019vv} proposed the voxel variational self encoder network VV-net, and the point distribution in each voxel is captured by the self encoder. In 2020, Shao \emph{et al.} \cite{shao2018h} proposed the data structure of opportunity space hash, designed hash2col and col2hash, so that CNN operations such as convolution and pooling can be parallelized.

\subsection{View-based Network.}
Usually, the point cloud is projected into the 2D image first, and then the 2D CNN is used to extract the image features. Due to the limitations of the existing deep learning network, this kind of method can only recognize the point cloud model from a specific angle. In 2017, Lawin \emph{et al.} \cite{lawin2017deep} generated images with different pitch angles and translation distances by controlling the equidistant angle. Snapnet-r proposed by Gueery \emph{et al.} \cite{guerry2017snapnet} can use 2D images and 3D as spatial structure information at the same time.  The mvpnet proposed by Jaritz \emph{et al.} \cite{jaritz2019multi} in 2019 can aggregate 2D image features into 3D. The relationship network proposed by Yang \emph{et al.} \cite{yang2019learning} comprehensively considers the relationship between different views and regions, and also uses the attention mechanism to generate scores to reflect the relative discrimination ability of views.

\begin{figure*}[t]
    \centering
   \includegraphics[width=18cm]{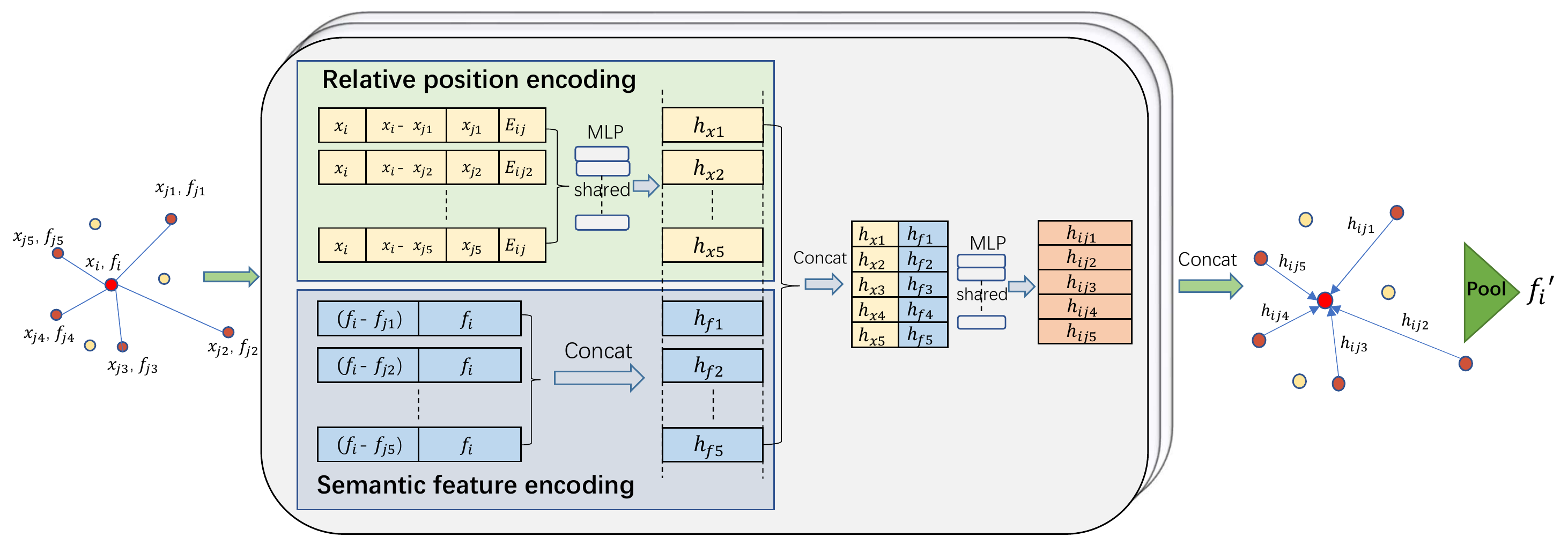}
   \caption{Illustration of feature extraction by DFA layer. The color closeness represents the adjacent points in the feature domain rather than the spatial neighbors. Rich information is obtained through relative position encoding and semantic feature encoding. The edge features of each adjacent point are obtained by sharing MLP, and finally the features of the central point are updated by maximum pooling operation. The subscript $j1 \cdots j5$ index the feature-domain neighbors for center $x_i$.}
\label{DFA}
\end{figure*}

\subsection{Point-based Network.}
Direct processing of point clouds contains complete original information. Qi \emph{et al.} \cite{qi2017pointnet} proposed Pointnet network, which is the first deep neural network to directly process disordered point clouds. Since it does not consider local features, they  \cite{qi2017pointnet++} further proposed Pointnet++ to extract local features at multiple levels. Later Atzmon \emph{et al.} \cite{atzmon2018point} proposed point convolution neural network, which uses expansion operator and constraint operator to generate convolution. In response to the problem of inflexibility of fixed grids, Thomas \emph{et al.} \cite{thomas2019kpconv} proposed KPconv, which is located in Euclidean space and is very effective in classifying point clouds with different densities. In addition, PointConv \cite{wu2019pointconv} and PointCNN \cite{li2018pointcnn} use 3D convolution kernels to extract features instead of sharing MLP. The PointConv \cite{wu2019pointconv} can be extended to deconvolution to achieve better segmentation results. And PointCNN \cite{li2018pointcnn} introduced the x-transform to rearrange the points into a potentially regular order, and then use convolution to extract local features from the point cloud.

\emph{Graph-based Methods.} By constructing a local or global graph structure to update delivery messages and learn features. In general, the graph structure of the spatial domain relies on finding k-nearest neighbors for message passing, and the graph structure of the spectral domain needs to be realized by methods such as Laplace matrix spectral decomposition and Chebyshev polynomial approximation. KCNet \cite{shen2018mining} defines a point set kernel as a set of learnable 3D points. It aggregates repetitive features at 3D locations on the nearest neighbor graph based on geometric relationships and local high-dimensional features measured by kernel correlations. Wang \emph{et al.} \cite{wang2019dynamic} proposed DGCNN to learn the embedding of edges by constructing local graphs. Unlike DGCNN \cite{wang2019dynamic}, 3DGCN \cite{lin2020convolution} defines learnable kernels using graph max pooling mechanism, and introduces shift invariance and scale invariance into deep learning networks. DeepGCNs \cite{li2019deepgcns} uses residual connections and dilated convolutions to train deeper graph structures, and experiments confirm the positive effect of depth.

\emph{Transformer-based Methods.} Since the great success of transformers in the NLP field, a lot of work has also introduced attention mechanisms to related tasks in point clouds recently. PCT \cite{guo2021pct} adopts a similar architecture to pointnet \cite{qi2017pointnet}, using neighbor information embedding, and improved offset transformer for feature learning, so that it has achieved good results in classification and segmentation tasks. Similarly, there are also some research works based on the pointnet++ \cite{qi2017pointnet++} network, such as PT \cite{zhao2021point} and BL-Net \cite{han2022blnet} . The PT \cite{zhao2021point} proposed by Zhao \emph{et al.} is to add a layer of transformer to extract features after each downsampling or upsampling. The transformer has been modified to measure the difference between the corresponding channels between two eigenvectors (Q and K). BL-Net \cite{han2022blnet} newly designed position feedback module to perform feature-guided point shifting. In addition, Yan \emph{et al.} \cite{yan2020pointasnl} also used the attention mechanism and proposed PointASNL that can effectively process point clouds with noise.

\section{Methodology}
Extracting and utilizing effective features is crucial in point cloud tasks. We construct a local graph structure through dynamic updating, and the information can diffuse nonlocally in the whole point cloud. Based on the graph structure, we explore both the latent location and semantic features of different layers. Further, we make full use of global features and local features containing detailed information. We describe the operation called Dynamic Feature Aggregation (DFA) in Section 3.1, and then the network structure is introduced in Section 3.2.

\begin{figure*}[t]
\begin{center}
   \includegraphics[width=17cm,height=6cm]{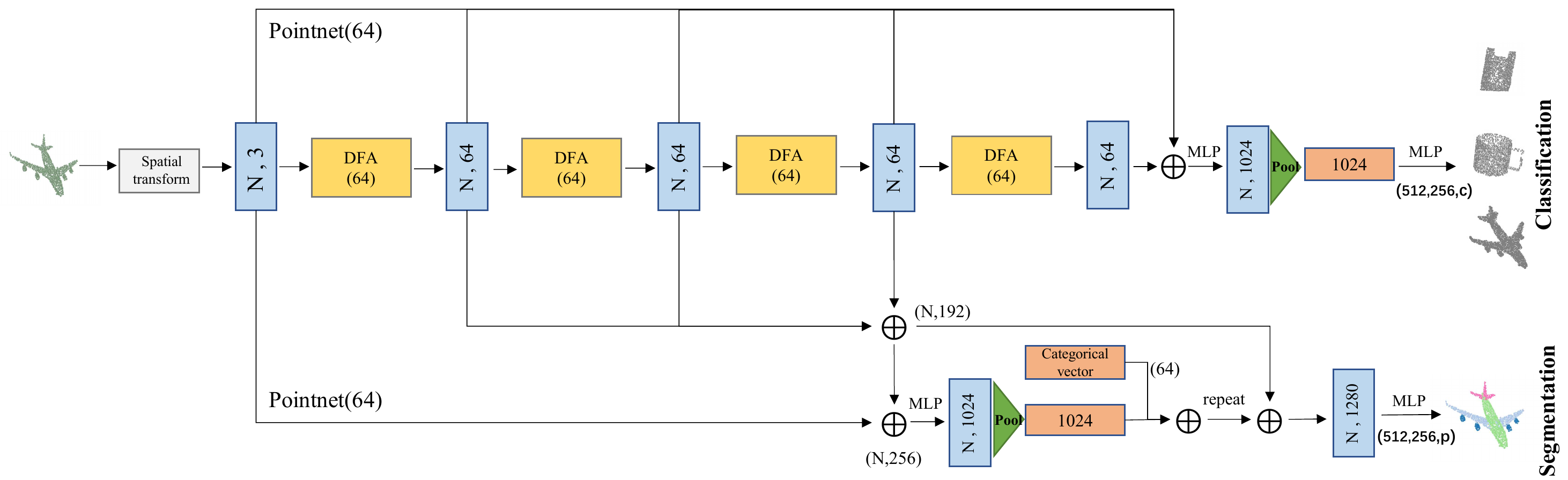}
\end{center}
   \caption{DFA based network architectures for classification and segmentation tasks. $\oplus$ stands for concatenated operations. The spatial transformation is designed to compute a $3\times 3$ matrix to align the input point cloud to the canonical space.  By concatenating local features and low-dimensional global features through MLP and max pooling, 1D global descriptors can be generated for classification tasks. For part segmentation, we  generate 1024-dimensional global features, fuse the category feature vectors, and then concatenate the detailed local features again to output the category score of each point through MLP.}
\label{architecture}
\end{figure*}

\subsection{Dynamic Feature Aggregation}
We define the input point cloud as $X=\left\{x_i|i=1,2,...,N\right\}\in\mathbb{R}^{N\times3}$ with the corresponding features defined as $F=\left\{f_i|i=1,2,...N\right\}\in\mathbb{R}^{N\times D}$. Here $x_i$ represents the three-dimensional coordinates $(x,y,z)$ of the {\emph i}-th point. As the input point cloud only contain three-dimensional coordinates, the geometry coordinates can also be regarded as its initial feature.

When extracting features at each layer, a local graph needs to be dynamically constructed, which is defined as $\mathcal{G}=(\mathcal{V},\mathcal{E})$, where $\mathcal{V}=\left\{1,2,...n\right\}$ and $\mathcal{E}\subseteq \mathcal{V}\times\mathcal{V}$ are the {\em vertices} and {\em edges}, respectively. We construct a local graph structure by finding k-nearest neighbors in the feature domain, including self-loops. Suppose that $x_i$ is the center point of the graph structure, and then $N(i)=\left\{j:(i,j)\in\mathcal{E}\right\}$ is the neighboring point in the feature domain. Specifically, the similarity of features is calculated and measured in the same way as Euclidean space distance in each feature dimension, and the k points with the smallest value are selected as the nearest neighbors. Then retrieve the 3D coordinates of each nearest neighbor. Given the input three-dimensional coordinates and D-dimensional features, our purpose is to learn and output M-dimensional features with the same number of points through the DFA layer.

Because we establish the connection between the center point and the surrounding k-nearest neighbors by building a local graph structure, so we define the feature of the edge as $e_{ij} = h_\Theta(f_i,f_j)$ , where $h_\Theta:\mathbb{R}^D \times \mathbb{R}^D \rightarrow \mathbb{R}^{M}$ is a nonlinear function with a set of learnable parameters $\Theta$. Finally, we aggregate the edge features of the k nearest neighbors along each channel, and obtain the result for each center point $f_i$ that enters the DFA layer feature extraction, which is defined as follows:
\begin{equation}
    f_i^{'} = \underset{j\in N(i)}{\Pi}h_\Theta(f_i,f_j)\label{overall}
\end{equation}

{\bf Semantic Feature Encoding.} We choose to find k-nearest neighbors in the feature domain, which means that the points sharing the same class will have high probability to be connected. Then we concatenate the feature of the center point and the feature differences with its neighbors as semantic feature information. Because this not only includes the features of all the original center points, but also transmits information to the surrounding points through the feature difference with the neighbors. And we define the encoding as follows:
\begin{equation}
h_{fj} =  f_i\oplus (f_i-f_{j}),j\in N(i)
\end{equation}
Here, $\oplus$ is the concatenate operation. We calculate and concatenate the feature differences and its own features along each dimension, aiming to encode semantically similar features and explore their latent information.

{\bf Relative Position Encoding.} We first need to store the original 3-dimensional position coordinate, and then find the latent position information of the corresponding nearest neighbors in the feature domain for each center point. We use the relative position information of the neighboring points to encode as follows:
\begin{equation}
h_{xj} = MLP (x_i\oplus x_{j}\oplus (x_i-x_j)\oplus \parallel x_i-x_{j}\parallel),j\in N(i)
\end{equation}
where $x_i$ and $x_{j}$ represent the original three-dimensional coordinates, $(x_i-x_j)$ calculate the relative coordinates of the center point and the k-nearest neighbors of the feature domain , $\oplus$ is the concatenate operation, and $\parallel \cdot \parallel$ calculates the Euclidean distance between the neighbours and center point. Unlike finding the  nearest neighbors in the  space restricted by geometry distance, we can discover more latent location information in the feature domain that may have similar semantic feature but with larger geometry distance.

When obtaining the position  and semantic embedding, we can concatenate these two parts  first and then extract the edge features through the MLP operation:
\begin{equation}
h_{ij} = MLP (h_{x_j}\oplus h_{f_j}),j\in N(i)
\end{equation}

Finally, we need to consider how to aggregate the features of the neighboring edges, that is {\em $\Pi$} in \eqref{overall}. We have three options for the over-aggregation $\Pi$. The first is to maximize the pool of edge features learned by all nearest neighbors to obtain the features of the center point. The second is to add all edge features. The third is to perform softmax on the neighbors to obtain a weight coefficient $W_{ij}$, and then multiply it with each edge feature, that is, $W_{ij}\times h_{ij}$ to obtain the attentive edge feature, and finally add and update the features of the center point. The experimental results show that the first maximum pooling has the best performance, so we choose the maximum pooling to aggregate all edge features.

\subsection{Network Architecture}
We use the proposed DFA layer to design two network architectures for the point cloud classification and segmentation task as shown in Fig. \ref{architecture}. We send the initial  point cloud into a spatial transformation network similar to the Pointnet \cite{qi2017pointnet} network. By learning the position information of the point cloud itself, we can learn a  rotation matrix that is most conducive to the classification or segmentation. The point clouds are multiplied and fed into our stacked DFA layer to extract features.

{\bf Local and Global Information Aggregation.} Focusing only on the global features obtained by pooling on each point ignores the local interaction between points. Or only focusing on local features of surrounding points is one-sided. Therefore, we choose a combination of local features and global features to comprehensively learn the information contained in the point cloud, so that it can be better used in classification and segmentation tasks. Our local features are learned by several layers of DFA, and the lower-dimensional global features is obtained similarly to Pointnet \cite{qi2017pointnet} by using shared MLP and max pooling. Our ablation experiments have also confirmed that integration with global feature is beneficial. On the other hand, we set several local features and low-dimensional global features to the same dimension (64) because we think they are equally important, which is also confirmed in practice.

{\bf Classification Network.} Our classification network is shown in the upper part of Fig. \ref{architecture}, and the point cloud through the spatial transformation network is sequentially passed through four DFA to extract local features. The input of each layer is the output of the previous layer. We concatenate these four local features and the global features extracted from the initial point cloud, and then convert them to higher dimensions through MLP operations. Finally, global features are obtained by max pooling for classification prediction.

{\bf Segmentation Network.} Our segmentation network is similar to the classification network, as shown in the lower part of Fig. 
 \ref{architecture}. We pass the transformed point cloud through three DFA layers in sequence. The three local features and low-dimensional global features are also concatenated to obtain a 1024-dimensional global features through MLP and max pooling. If it is part segmentation, then we add a category feature vector (64). If it is semantic segmentation, it will not be added. Finally we use the shared MLP to resize the features and predict the semantic label for each point.

{\bf Dynamic Graph Update.} Depending on the spatial interaction of the point cloud, locally adjacent parts can form subsets. However, considering the spatial neighbors for graph update sometimes leads to failure of feature aggregation.  For example, for the point clouds of air plane,  the aircraft wing and fuselage are adjacent in space, the mutually updated features are useless. So we use the point of finding k-nearest neighbors on the feature domain, which means that these points can constitute meaningful parts. Each time we find neighbors in the feature domain to reconstruct the local graph structure. It can be said that our graph is dynamically updated, so we can explore more latent location information, which is also a limitation that cannot be achieved by doing k-nearest neighbors in space.

\section{Experiments}
In this section, we evaluate our models using DFA for point cloud classification and part segmentation tasks. 

\begin{table}[h]
\begin{center}
\scalebox{0.85}{\begin{tabular}{l c c c c}
\hline
Methods &Input & point & mAcc & OA\\
\hline
Pointnet\cite{qi2017pointnet}&xyz & 1k & 86.0 & 89.2 \\
Pointnet++\cite{qi2017pointnet++}&xyz & 1k & - & 90.7 \\
Pointnet++\cite{qi2017pointnet++}&xyz,normal & 5k & - & 91.9 \\
SpiderCNN\cite{xu2018spidercnn}&xyz,normal & 1k & - & 92.4 \\
PointWeb\cite{zhao2019pointweb}&xyz,normal & 1k & 89.4 & 92.3 \\
PointCNN\cite{li2018pointcnn}&xyz & 1k & 88.1 & 92.2\\
DGCNN\cite{wang2019dynamic}&xyz & 1k & 90.2 & 92.2\\
Point2Sequence\cite{liu2019point2sequence}&xyz & 1k & 90.4 & 92.6\\
FPConv\cite{lin2020fpconv}&xyz,normal & 1k  & - & 92.5\\
PointConv\cite{wu2019pointconv}&xyz,normal & 1k & - &92.5\\
KPConv\cite{thomas2019kpconv}&xyz & 6k  & - & 92.9\\
Point2Node \cite{han2020point2node}&xyz & 1k & - & 93.0\\
PointASNL\cite{yan2020pointasnl}&xyz & 1k & - &92.9\\
PointASNL\cite{yan2020pointasnl}&xyz,normal & 1k & - &93.2\\
PCT\cite{guo2021pct}&xyz & 1k & - &93.2\\
SO-Net\cite{li2018so}&xyz,normal & 5k & 90.8 & 93.4 \\
BL-Net\cite{han2022blnet} &xyz & 1k&- &93.5\\
AG-conv\cite{zhou2021adaptive} &xyz & 1k&90.7 &93.4\\
PointStack\cite{wijaya2022advanced}&xyz & 1k & 89.6 & 93.3\\
\hline
Ours(1024 points) &xyz & 1k& \textbf{91.1} & \textbf{93.6}\\
Ours(2048 points)&xyz & 2k & \textbf{91.6} & \textbf{94.0}\\
\hline
\end{tabular}}
\end{center}
\caption{Classification results on ModelNet40.}
\label{classification result}
\end{table}

\begin{figure*}[t]
\centering 
\includegraphics[height=9cm,width=16.5cm]{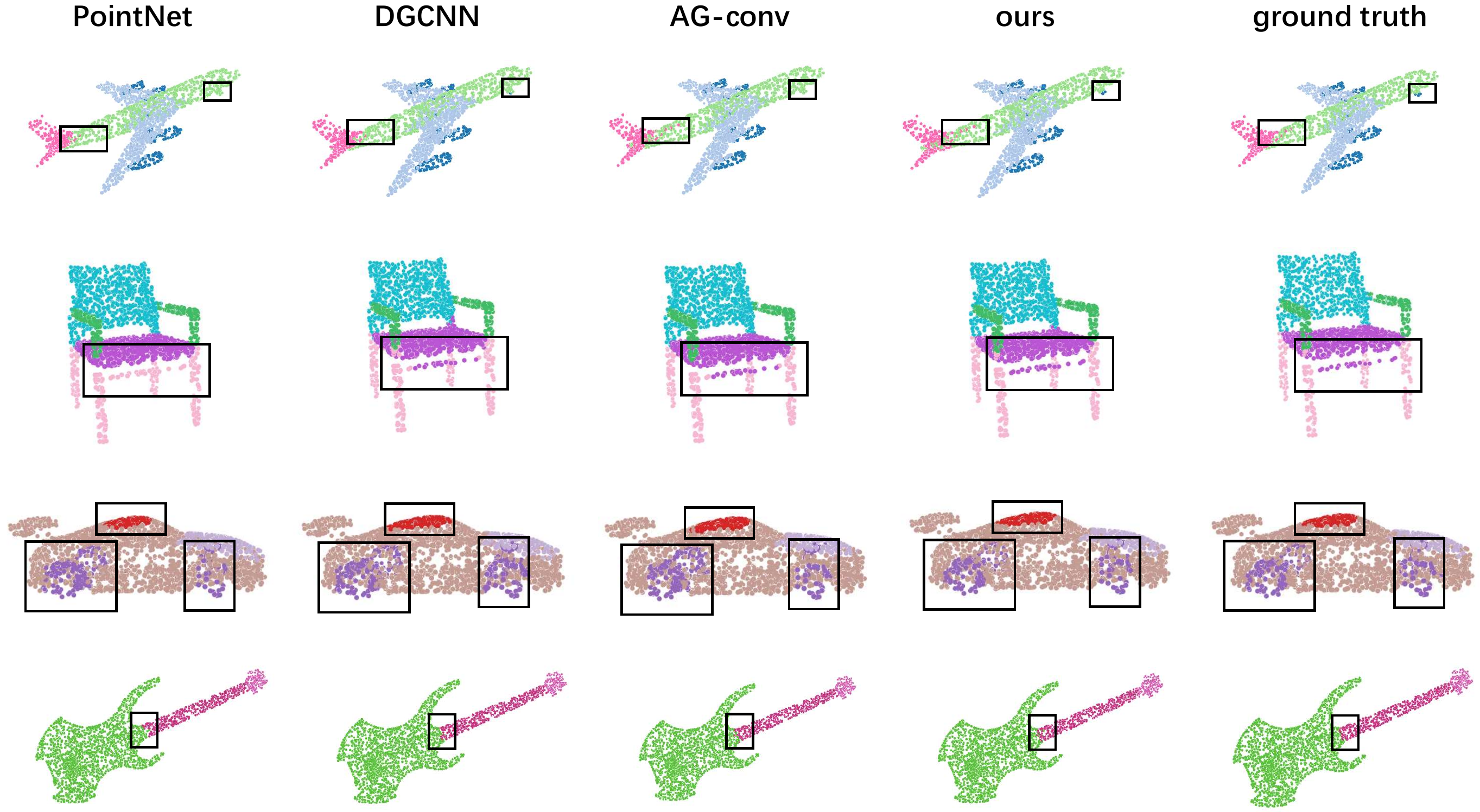}
\caption{Visual comparison of four methods for part segmentation. }
\label{visual}
\end{figure*}

\begin{table*}[t]
\begin{center}
\resizebox{1\textwidth}{18mm}
{
\begin{tabular}{c|c|c|c|c|c|c|c|c|c|c|c|c|c|c|c|c|c}
\hline
Methods& mIou & air. & bag & cap  & car & cha. & ear.  & gui. & kni. & lam. & lap. & mot. & mug & pis. & roc. & ska.  & tab.\\
\hline
NUM &  & 2690 & 76 & 55 & 898 & 3758 & 69 & 787 & 392 & 1547 & 451 & 202 & 184 & 283 & 66 & 152 & 5271 \\
\hline
Pointnet\cite{qi2017pointnet} & 83.7 & 83.4 &78.7 & 82.5 & 74.9 & 89.6 & 73.0 & 91.5 & 85.9 & 80.8 & 95.3 &65.2 & 93.0 & 81.2 & 57.9 & 72.8 & 80.6 \\
Pointnet++\cite{qi2017pointnet++} & 85.1 & 82.4 &79.0 & 87.7 & 77.3 & 90.8 & 71.8 & 91.0 & 85.9 & 83.7 & 95.3 &71.6 & 94.1 & 81.3 & 58.7 & 76.4 & 82.6 \\
SO-Net\cite{li2018so} & 84.9 & 82.8 &77.8 & 88.0 & 77.3 & 90.6 & 73.5 & 90.7 & 83.9 & 82.8 & 94.8 &69.1 & 94.2 & 80.9 & 53.1 & 72.9 & 83.0\\
RGCNN\cite{te2018rgcnn} & 84.3 & 80.2 &82.8 & \textbf{92.6} & 75.3 & 89.2 & 73.7 & 91.3 & \textbf{88.4} & 83.3 & 96.0 &63.9 & \textbf{95.7} & 60.9 & 44.6 & 72.9 & 80.4\\
DGCNN\cite{wang2019dynamic} & 85.2 & 84.0 &83.4 & 86.7 & 77.8 & 90.6 & 74.7 & 91.2 & 87.5 & 82.8 & 95.7 &66.3 & 94.9 & 81.1 & \textbf{63.5} & 74.5 & 82.6\\
PCNN\cite{atzmon2018point} & 85.1 & 82.4 &80.1 & 85.5 & 79.5 & 90.8 & 73.2 & 91.3 & 86.0 & 85.0 & \textbf{96.7} &73.2 & 94.8 & 83.3 & 51.0 & 75.0 & 81.8\\
3D-GCN\cite{lin2020convolution} & 85.1 & 83.1 & 84.0 & 86.6 & 77.5 & 90.3 & 74.1 & 90.9 & 86.4 & 83.8 & 95.3 &65.2 & 93.0 & 81.2 & 59.6 & 75.7 & 82.8\\
PointASNL\cite{yan2020pointasnl} & 86.1 & 84.1 &84.7 & 87.9 & 79.7 & 92.2 & 73.7 & 91.0 & 87.2 & 84.2 & 95.8 &74.4 & 95.2 & 81.0 & 63.0 & \textbf{76.3} & 83.2\\
PRA-Net\cite{2021PRA} & 86.3 & 84.4 &\textbf{86.8} & 89.5 & 78.4 & 91.4 & \textbf{76.4} & 91.5 & 88.2 & \textbf{85.3} & 95.7 &73.4 & 94.8 & 82.1 & 62.3 & 75.5 & \textbf{84.0}\\
\hline
Ours & 86.0 & \textbf{85.4} &80.0 & 85.8 & \textbf{80.6} & \textbf{92.4} & 74.1 & \textbf{92.0} & 87.4 & 84.6 &95.6 & \textbf{73.5} & 94.4 & \textbf{83.9} & 59.0 & 74.0 &83.2\\
\hline
\end{tabular}}
\end{center}
\caption{Part segmentation results on ShapeNet dataset. Metric is mIoU(\%).}
\label{segmentation result}
\end{table*}

\subsection{Classification}
{\bf Data.} We evaluate our point cloud classification model on the ModelNet40 \cite{wu20153d} dataset. This dataset contains 12311 mesh CAD models from 40 categories, where 9843 models are used for training and 2468 models are used for testing. We follow the experimental setting of \cite{qi2017pointnet}. We uniformly sample 1024 or 2048 points for each model, each using only 3D coordinates $(x,y,z)$ as input. Data augmentation operations include point shifting, scaling and perturbing of the points. 

{\bf Network Configuration.} The network architecture is shown in Fig. \ref{architecture}. At each layer we recompute the graph based on feature similarity. For the 1024 points we set the number of nearest neighbors k value to 20, and to maintain the same density, we set  k  to 40 for the 2048 points. We use four DFA layers to extract local geometric features and a Pointnet-like structure to extract low-dimensional global features. These are implemented using fully connected layers (64). We connect the extracted multi-layer features to obtain $64\times5=320$-dimensional features. Then the global features are obtained, and then two fully connected layers are used to transform the global features for classification. All layers use LeakyReLU and batch normalization. We use the SGD optimizer with momentum of 0.9. The initial learning rate is 0.1, and the random drop rate of the fully connected layer is 0.5 to prevent overfitting. The batch size is set to 32. We use Pytorch implementation and train the network on two RTX 2080Ti GPUs.

{\bf Results.} Table \ref{classification result} shows the results of the classification task, and the evaluation metrics we use on this dataset are the average class accuracy and overall accuracy. Our network only feeds 3D coordinates into training, which contains less raw information, but achieves the best results on this dataset. The test result of 2048 sampling points is better than that of 1024 points, indicating that when more original information is included, our network can learn more features and have better performance.

\begin{table*}[t]
\begin{center}
\resizebox{1\textwidth}{18mm}
{
\begin{tabular}{c|cc|ccccccccccccc}
\hline
Methods&mAcc& mIou & ceiling & floor & wall  & beam & column & windows  & door & chair & table & bookcase & sofa & board & clutter \\
\hline
Pointnet\cite{qi2017pointnet} & 48.98 & 41.09 &88.80 & 97.33 & 69.80 & 0.05 & 3.92 & 46.26 & 10.76 & 58.93 & 52.61 &5.85 & 40.28 & 26.38 & 33.22  \\
SEGCloud\cite{tchapmi2017segcloud} & 57.35 & 48.92 &90.06 & 96.05 & 69.86 & 0.00 & 18.37 & 38.35 & 23.12 & 70.40 & 75.89 &40.88 & 58.42 & 12.96 & 41.60  \\
PointCNN\cite{li2018pointcnn} & 63.86 & 57.26 &92.31 & 98.24 & 79.41 & 0.00 & 17.60 & 22.77 & 62.09 & 74.39 & 80.59 &31.67 & 66.67 & 62.05 & 56.74  \\
PointWeb\cite{zhao2019pointweb} & 66.64 & 60.28 &91.95 & 98.48 & 79.39 & 0.00 & 21.11 & 59.72 & 34.81 & 76.33 & \textbf{88.27} &46.89 & \textbf{69.30} & 64.91 & 52.46  \\
SPG\cite{landrieu2018large} & 66.50 & 58.04 &89.35 & 96.87 & 78.12 & 0.00 & \textbf{42.81} & 48.93 & 61.58 & \textbf{84.66} & 75.41 &69.84  & 52.60 & 2.10 & 52.22  \\
PCNN\cite{atzmon2018point} & 67.01 & 58.27 &92.26 & 96.20 & 75.89 & \textbf{0.27} & 5.98 & 69.49 & 63.45 & 66.87 &65.63 & 47.28 & 68.91 & 59.10 & 46.22 \\
PCT\cite{guo2021pct} & 67.65 & 61.33 &92.54 & 98.42 & \textbf{80.63} & 0.00 & 19.35 & 61.64 & 48.00 & 76.58 &85.20 & 46.22 & 67.71 & \textbf{67.93} & 52.29 \\
\hline
Ours & \textbf{67.96} & \textbf{62.18} &\textbf{92.68} & \textbf{98.50} & 79.12 & 0.05 & 36.72 & \textbf{67.45} & \textbf{65.18} & 75.36 &86.77 & \textbf{71.52} & 52.59 &65.02& \textbf{57.12}\\
\hline
\end{tabular}}
\end{center}
\caption{Semantic segmentation results on S3DIS dataset.}
\label{3D segmentation result}
\end{table*}

\begin{figure*}
\centering 
\includegraphics[height=9cm,width=16.5cm]{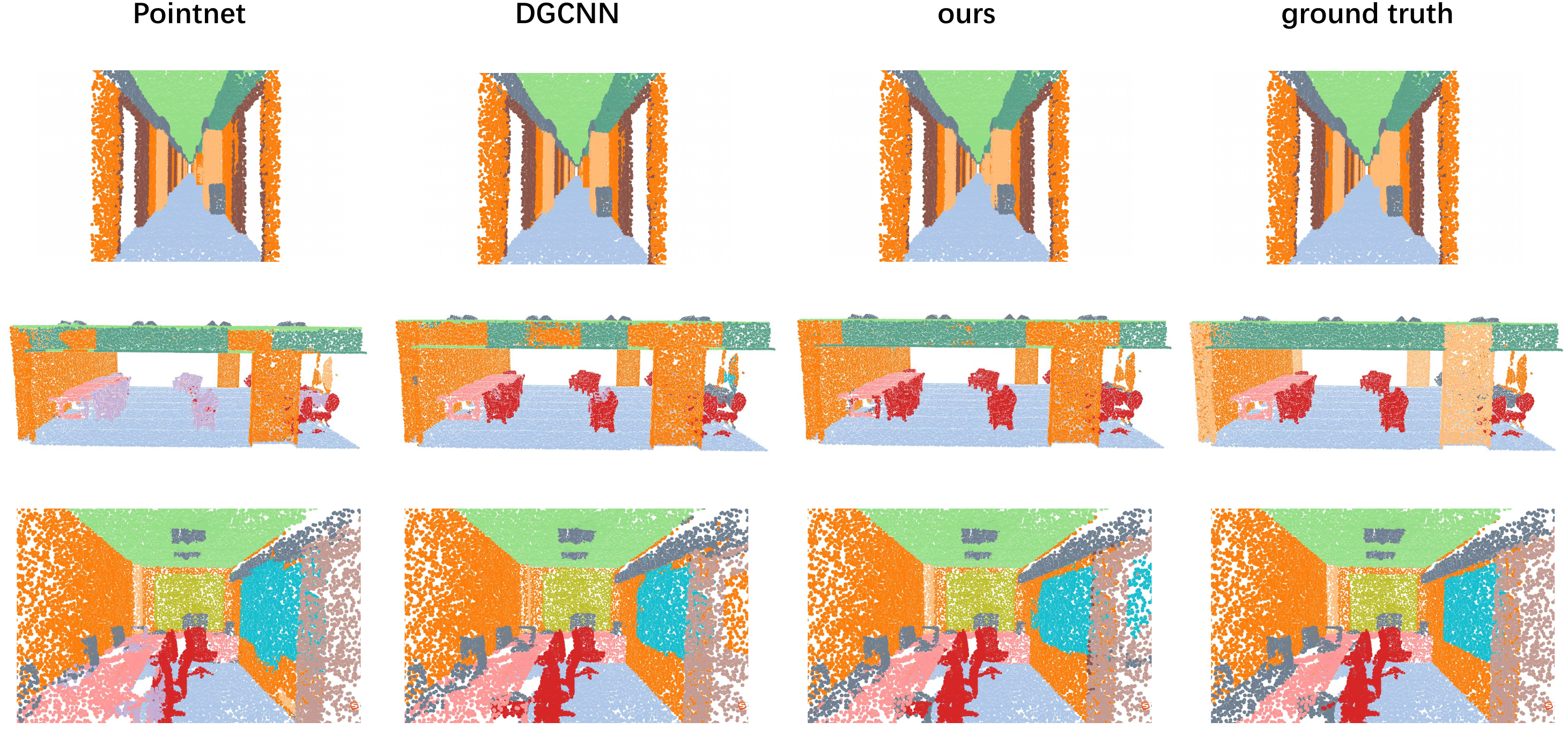}
\caption{Visual comparison of three methods for semantic segmentation. }
\label{sem_visual}
\end{figure*}

\subsection{Part Segmentation}
{\bf Data.} We test our model on the ShapeNet dataset \cite{yi2016scalable} for point cloud part segmentation. This dataset contains 16881 shapes in 16 categories, of which 14006 are used for training and 2874 are used for testing. There are 50 parts tags in total, and each model includes 2-6 parts. We follow the experimental setup of \cite{qi2017pointnet}. 2048 points are sampled from each shape, and the input consists only of the 3D coordinates.

{\bf Network Configuration.} We use three DFA layers to extract features, and operate the same as classification to obtain 1024-dimensional global features. Following \cite{wang2019dynamic}, we also add a one-hot vector representing the category type to each point. Then we concatenate global features and category vectors as new global features with $1024+64=1088$-dimensions. We re-concatenate the previous three local features and convert them into the features of each point through three fully connected layers (512, 256, 128) for segmentation. The settings of our training parameters are the same as in the classification task, except that the batch size is changed to 16.

{\bf Results.} We evaluate the performance of part segmentation by the mIou metric. The Iou of a shape is computed by averaging of each part. The mean Iou (mIou) is calculated by averaging the Ious of all testing instances. From the experimental results in table \ref{segmentation result}, it can be seen that in some categories with a small number of samples, the segmentation effect is not good due to too few training samples. But overall, our method has better performance, especially with the highest mIou in many categories such as airplane, car, chair, etc. This benefits from these categories having sufficient samples so that our network can learn rich features for part segmentation tasks. Fig. \ref{visual} shows the visual differences between us and several other mainstream methods on some categories. These methods are roughly capable of distinguishing different parts of an object, and the difference lies in the identification of details. Looking closely at the tail section of the airplane, the fence section below the chair, the top of the car, and the connection between different parts in the guitar, our method is closer to the ground truth.

\subsection{Semantic Segmentation}

{\bf Data.} We further test our model on the Stanford Large-Scale 3D Indoor
Spaces Dataset (S3DIS) dataset \cite{armeni20163d} for point cloud semantic scene segmentation. This dataset is taken from 271 rooms in 6 different areas in 3 different buildings. The point cloud data of each scene has 9-dimensional data including $xyz$ three-dimensional coordinates, RGB color information, and the normalized position coordinates $x'y'z'$ of each point relative to the room where it is located. At the same time, each point cloud in the scene is assigned a semantic label from 13 categories (such as ceiling, table, etc.). 

{\bf Network Configuration.} Our semantic segmentation network configuration is the same as for part segmentation, the only difference is that no feature vector is added.

{\bf Results.} We divide each room into 1m × 1m blocks and sample 4096 points in each block during training. And we use area5 as the test set. For evaluation metrics, we use mean class accuracy (mAcc) and mean class intersection (mIou). The experimental results are shown in the table \ref{3D segmentation result}, and the visualization is shown in the fig. \ref{sem_visual}.

\subsection{Ablation Studies}

In this subsection, we explore the effect of using different choices in the network. The effectiveness of our module and parameter selection is demonstrated in these ablation experiments.

{\bf Number of neighbors.} The k value of constructing the local graph structure has a great influence on the extracted features. Therefore, it is very important to choose an appropriate value of k in the experiment. We conducted 4 sets of experiments to explore the impact of choosing different k values on the classification results of 2048 points, which is also shown in the table \ref{k}. When the value of k is 10 and 20, the neighborhood of each center point is small and cannot fully interact with the neighbor points. Appropriately increasing the value of k can also have room for improvement, which also shows that DFA can effectively use the features of neighborhood points to learn local features. By further increasing the value of k, it can be found that increasing the value of k all the time will not increase the accuracy of the model. Because when the value of k is too large, there will be many noise points that are very different from the center point features, which is useless or even burdensome for updating the center point features, and will also increase the amount of parameters and network training time. Choosing a neighbor k value of 40 can obtain the best average class accuracy and overall accuracy.

\begin{table}[h]
\begin{center}
\begin{tabular}{l|c c}
\hline
 k  & mAcc & OA\\
\hline
10 & 90.2 & 93.3 \\
20 & 90.8 & 93.7 \\
40 & 91.6 & 94.0\\
60 & 91.5 & 93.3\\
\hline
\end{tabular}
\end{center}
\caption{Number of neighbors(k)}
\label{k}
\end{table}

{\bf Selection of aggregate functions $\Pi$.} It can be seen in many previous works\cite{qi2017pointnet}\cite{qi2017pointnet++}\cite{guo2021pct} that some symmetric pooling functions such as max/sum/mean are often used to overcome the disordered characteristics of point clouds. In our DFA layer, we also need to aggregate edge features to update features for each center point. We experimented with different aggregation functions such as max, sum, or sum with attention weights which first do softmax on k-nearest neighbors dimension to get the attention weights and then multiply and accumulate them accordingly. The max function is to select the largest feature of points in the local neighborhood. The sum function is to add the features of all points in the neighborhood, and the mean function is to divide by the k value after the sum function. Table \ref{aggregate functions} shows the results of our selection of different aggregation functions on a classification experiment of 2048  points. Although the maximum pooling function will lose the non-largest part of the features, it will retain the largest part of the most significant features, and the experimental results show that it is the most effective. We finally choose the best-performing max function to aggregate the edge features.

\begin{table}[h]
\begin{center}
\begin{tabular}{l|cc}
\hline
 $\Pi$  & mAcc & OA\\
\hline
max & 91.6 & 94.0 \\
sum & 90.5 & 93.4 \\
mean & 90.3 & 93.2 \\
attention sum & 91.0 & 93.5 \\
\hline
\end{tabular}
\end{center}
\caption{Choice of different aggregation functions $\Pi$}
\label{aggregate functions}
\end{table}

{\bf Feature or space domains.} Further, we explore in which domain is better to compute k-nearest neighbors, i.e., the feature domain or  the spatial domain. If we choose to do k-nearest neighbors in the spatial domain, it means that the graph structure is fixed each time. On the one hand, the relative position coding will be the same, on the other hand, it is very limited to exchange information with fixed neighbor points each time. If we choose to do k-nearest neighbors on the feature domain, it means that the local graph structure is dynamically updated, and the neighbors of the graph are different each time but the features are similar. We can make better use of DFA layers to discover efficient features. We choose to compare the experimental results in the classification task of 2048 points. As can be seen from the table \ref{whether_spatial}, our way of exchanging information with neighbor updates in the feature domain is better. Because the k-nearest neighbors obtained in this way are more homogeneous. Especially for part segmentation, spatially adjacent points are not necessarily of the same class, so it is useless or even redundant to exchange information with these points.

\begin{table}[h]
\begin{center}
\begin{tabular}{c|c c}
\hline
 spatial or feature domain & mAcc & OA\\
\hline
feature & 91.6 & 94.0 \\
spatial & 91.1 & 93.4 \\
\hline
\end{tabular}
\end{center}
\caption{Comparison of k-nearest neighbors in feature domain and space.}
\label{whether_spatial}
\end{table}

{\bf Relative position information.} By computing the k-nearest neighbors of the feature domain, we are able to discover latent-location feature information that is not limited by space. In this way, the relative position encoding in each DFA layer is different because the neighborhood points are changing. This allows us to connect points that may not be in close spatial locations. So we explore its effectiveness by whether incorporating this part in the classification task of 2048 points. The experimental results in table \ref{position} show that adding location information encoding can have better performance. This also shows that the potential position information obtained by relative position encoding is crucial.

 \begin{table}[h]
\begin{center}
\begin{tabular}{l|cc}
\hline
 Position information  & mAcc & OA\\
\hline
w & 91.6 & 94.0 \\
w/o & 90.1 & 93.3 \\
\hline
\end{tabular}
\end{center}
\caption{Whether to add position information}
\label{position}
\end{table}

{\bf Low-dimensional global features.} Inspired by Pointnet \cite{qi2017pointnet} and Pointnet++ \cite{qi2017pointnet++}, it is not advisable to only focus on global features or local features, so we adopt a fusion of both. Global features can provide overall direction control, while local features can provide more detailed information. We believe that these are equally important in network learning, so after extracting local features of different depths, we concatenate these local features and low-dimensional global features together through MLP operations to upgrade to high-dimensional for subsequent tasks. To this end, we compare the classification results of 2048 points with or without adding low-dimensional global features. The table \ref{low_global} confirms the effectiveness of our way of concatenating the learned local features and low-dimensional global features.

\begin{table}[h]
\begin{center}
\begin{tabular}{l|cc}
\hline
 Low-global features & mAcc & OA\\
\hline
w & 91.6 & 94.0 \\
w/o & 89.9 & 93.1 \\
\hline
\end{tabular}
\end{center}
\caption{Whether to add low-dimensional global features}
\label{low_global}
\end{table}

\subsection{Model Complexity} We use the stat package in pytorch to output some quantitative results of the network model. It includes the total number of parameters of the network model, the number of floating-point operations required for network operation, and the memory occupied by node inference. The experimental results are all tested based on the classification model on 1024 points. At the same time, we test other mainstream methods for comparison as shown in the following table \ref{evaluation}.

It can be seen that our model has fewer parameters and does not occupy a large amount of memory, indicating that our network structure is lightweight, and not complicated and easy to implement. In networks based on graph methods, the amount of computation is generally too large due to the need to interact with neighbors to update features. Compared with other methods of this type, our floating-point operations are also much less. At the same time the performance is still the best.
\begin{table}[h]
\begin{center}
\begin{tabular}{c|cccc}
\hline
Method & Pparams & Flops & Memory & OA\\
\hline
Pointnet\cite{qi2017pointnet} & 0.7M & 0.5M & 10.5M & 89.2\\
Pointnet++\cite{qi2017pointnet++} & 2.2M & 3.1M & 231.5M & 91.9\\
DGCNN\cite{wang2019dynamic} & 1.8M & 1.89G & 123.0M & 92.9\\
AG-conv\cite{zhou2021adaptive} & 1.9M & 2.9G & 202.0M & 93.4\\
PCT\cite{guo2021pct} & 2.9M & 2.32G & 187.6M & 93.2\\
ours & 1.1M & 2.17G & 154.5M & 93.6\\
\hline
\end{tabular}
\end{center}
\caption{Quantitative evaluation of classification on ModelNet40.}
\label{evaluation}
\end{table}

\section{Conclusion} This paper proposes a new operation for point cloud learning and also demonstrates its performance in different tasks. The main contribution of our method is to aggregate local feature in the feature domain,  explore the latent relative position information and semantic feature information, and learn to obtain higher-dimensional features by concatenating local features and low-dimensional global features. Our DFA can dynamically construct graphs that are not spatially correlated and exchange information between points with semantically similar features. Experimental results show that our network outperforms the state-of-the-art on several public datasets. Further, our DFA module is simple and efficient, and can be seamlessly integrated into other network models.


\bibliography{mybibfile}

\end{document}